%% file: root.tex
\documentclass[letterpaper, 10 pt, journal, twoside]{ieeetran}
\usepackage{times}
\usepackage{epsfig}
\usepackage{graphicx}
\usepackage{amsmath}
\usepackage{amssymb}
\usepackage[percent]{overpic}
\usepackage{pifont}%
\usepackage{wasysym}

\usepackage[utf8]{inputenc} %
\usepackage[T1]{fontenc}    %
\usepackage{xcolor}         %
\definecolor{mypink1}{rgb}{0.78, 0.21, 0.13}
\usepackage[colorlinks=true]{hyperref}       %
\hypersetup{
    colorlinks=true,
    linkcolor=mypink1,
    filecolor=magenta,      
    }
\usepackage{url}            %
\usepackage{amsfonts}       %
\usepackage{nicefrac}       %
\usepackage{microtype}      %
\usepackage{tabularx}
\usepackage{booktabs}

\usepackage{graphicx}
\usepackage[export]{adjustbox}
\usepackage{xcolor}
\usepackage{overpic}
\usepackage{multirow}

\usepackage{times}
\usepackage{epsfig}
\usepackage{graphicx}
\usepackage{amsmath}
\usepackage{amssymb}
\usepackage[percent]{overpic}
\usepackage{pifont}%
\usepackage{wasysym}
\usepackage{booktabs}

\input{macros.tex}

\title{\LARGE \bf
SLAG: Scalable Language-Augmented Gaussian Splatting
}

\author{
Laszlo Szilagyi$^{1}$,
Francis Engelmann$^{1}$,
Jeannette Bohg$^{1}$ 
\thanks{$^{1}$
All authors are with the Department of Computer Science,
Stanford University, Stanford, CA 94309 USA.
}
\\[-5.0ex]%
\thanks{Manuscript received: December 6, 2024; Revised March 10, 2025; Accepted May 3, 2025.}%Use only for final RAL version
\thanks{This paper was recommended for publication by Editor Sven Behnke upon evaluation of the Associate Editor and Reviewers' comments. This work was supported in part by an SNSF PostDoc.Mobility Fellowship during Francis Engelmann’s stay at Stanford University.
} %Use only for final RAL version
\thanks{Digital Object Identifier (DOI): 10.1109/LRA.2025.3573203}
}

\begin{document}

\maketitle

\input{sections/01_abstract}
\input{sections/02_intro}

\input{sections/03_related_work}

\input{sections/04_method}
\input{sections/05_experiments}
\input{sections/06_conclusion}

{\small
\bibliographystyle{IEEEtran}
% \balance
\bibliography{egbib}
}

\end{document}

%% file: macros.tex
\usepackage{booktabs}
\usepackage{multirow}
\usepackage{overpic}
\usepackage{cite}
\usepackage[font=small,labelfont=bf]{caption}
\usepackage{color}
\usepackage{pifont}
\makeatletter
\@namedef{ver@everyshi.sty}{}
\makeatother
\usepackage{tikz}
\usetikzlibrary{patterns}

\usepackage{colortbl}
\usepackage{pgfplots}
\pgfplotsset{compat=1.17}
\usepackage{tabularx}
\usepackage{arydshln}
\usepackage{amssymb}%
\usepackage{pifont}%

\usepackage{wasysym}%

\usepackage{enumitem}
\usepackage{wrapfig}
\usepackage{xspace}
\usepackage{tabu}
\usepackage[super]{nth}
\usepackage{scalefnt}
\usepackage{dsfont}
\usepackage{graphicx,calc}
\usepackage{subcaption}
\usepackage{algorithm}
\usepackage{algpseudocode}

\definecolor{codegreen}{rgb}{0,0.6,0}
\definecolor{codegray}{rgb}{0.5,0.5,0.5}
\definecolor{codepurple}{rgb}{0.58,0,0.82}
\definecolor{backcolour}{rgb}{0.95,0.95,0.92}

\definecolor{ourPurple}{HTML}{9673A6}
\definecolor{ourOrange}{HTML}{D79B00}
\definecolor{ourGreen}{HTML}{82B366}
\definecolor{ourRed}{HTML}{B85450}
\definecolor{personColor}{HTML}{0000FF}
\definecolor{bgColor}{HTML}{bed4f3}

\definecolor{darkgreen}{RGB}{0,153,51}
\definecolor{linkgreen}{RGB}{52,130,48}
\definecolor{LightCyan}{rgb}{0.87,0.92,0.96}
\definecolor{m_green}{RGB}{233, 254, 187}
\definecolor{m_orange}{RGB}{255, 212, 121}
\definecolor{m_red}{RGB}{255, 190, 188}
\definecolor{m_violet}{RGB}{215, 131, 255}
\definecolor{m_blue}{RGB}{186, 234, 255}
\definecolor{m_brown}{RGB}{255,212,120}
\definecolor{m_lightgreen}{RGB}{212,251,122}
\definecolor{notetext}{rgb}{0.7,0,0}

\definecolor{model_pink}{RGB}{235, 106, 164}
\definecolor{model_orange}{RGB}{250, 194, 122}
\definecolor{model_green}{RGB}{164, 210, 162}
\definecolor{model_gray}{RGB}{120, 120, 120}
\definecolor{model_yellow}{RGB}{251, 231, 171}
\definecolor{model_purple}{RGB}{190, 146, 211}

\usepackage{amssymb}%
\usepackage{pifont}%

\newcolumntype{Y}{>{\centering\arraybackslash}X}
\newcolumntype{Z}{>{\raggedleft\arraybackslash}X}

\usepackage{array}
\newcolumntype{P}[1]{>{\centering\arraybackslash}p{#1}}
\newcolumntype{M}[1]{>{\centering\arraybackslash}m{#1}}

\definecolor{darkblue}{RGB}{60, 82, 145}
\definecolor{kingblue}{RGB}{65, 105, 225}

\definecolor{background}{RGB}{226, 226, 226}
\definecolor{head}{RGB}{210, 78, 142}
\definecolor{rightArm}{RGB}{255, 176, 0}
\definecolor{leftArm}{RGB}{228, 162, 227}
\definecolor{rightForeArm}{RGB}{90, 64, 210}
\definecolor{leftForeArm}{RGB}{243, 232, 88}
\definecolor{rightHand}{RGB}{158, 143, 20}
\definecolor{leftHand}{RGB}{192, 100, 119}
\definecolor{torso}{RGB}{100, 143, 255}
\definecolor{hips}{RGB}{129, 103, 106}
\definecolor{rightUpLeg}{RGB}{243, 115, 68}
\definecolor{leftUpLeg}{RGB}{152, 200, 156}
\definecolor{rightLeg}{RGB}{149, 192, 228}
\definecolor{leftLeg}{RGB}{152, 78, 163}
\definecolor{rightFoot}{RGB}{129, 0, 50}
\definecolor{leftFoot}{RGB}{76, 134, 26}

\newlength\myheight
\newlength\mydepth
\settototalheight\myheight{Xygp}
\usepackage{tikz}

\usepackage{caption} 

%% file: sections/01_abstract.tex
\begin{abstract}
Language-augmented scene representations hold great promise for large-scale robotics applications such as search-and-rescue, smart cities, and mining. Many of these scenarios are time-sensitive, requiring rapid scene encoding while also being data-intensive, necessitating scalable solutions. Deploying these representations on robots with limited computational resources further adds to the challenge.
To address this, we introduce SLAG, a multi-GPU framework for language-augmented Gaussian splatting that enhances the speed and scalability of embedding large scenes. Our method integrates 2D visual-language model features into 3D scenes using SAM~\cite{kirillov2023segany} and CLIP~\cite{radford2021learningtransferablevisualmodels}. Unlike prior approaches, SLAG eliminates the need for a loss function to compute per-Gaussian language embeddings. Instead, it derives embeddings from 3D Gaussian scene parameters via a normalized weighted average, enabling highly parallelized scene encoding. Additionally, we introduce a vector database for efficient embedding storage and retrieval.
Our experiments show that SLAG achieves an 18× speedup in embedding computation on a 16-GPU setup compared to OpenGaussian~\cite{wu2024opengaussian}, while preserving embedding quality on the ScanNet~\cite{dai2017scannet} and LERF~\cite{lerf2023} datasets. For more details, visit our project website: \url{https://slag-project.github.io/}.
\end{abstract}


%% file: sections/02_intro.tex
\section{Introduction} \label{sec:intro}

\IEEEPARstart{R}{ecent} progress in 3D scene representations \cite{lerf2023, engelmann2024opennerf, qin2023langsplat, shi2023language, wu2024opengaussian, yue2024improving} enhances 3D reconstruction techniques such as NeRF \cite{mildenhall2020nerf} and Gaussian Splatting \cite{kerbl3Dgaussians} by integrating language embeddings from 2D vision-language models like CLIP \cite{radford2021learningtransferablevisualmodels}.

Unlike traditional 3D segmentation methods \cite{wu2024ptv3, takmaz20233d, yue2023agile3d},
these approaches typically reconstruct
the scene using NeRF or Gaussian Splatting, extract language-
aligned embeddings, and enable querying the scene based on natural
language.
Methods like LERF~\cite{lerf2023}, OpenNeRF~\cite{engelmann2024opennerf}, LangSplat \cite{qin2023langsplat}, and OpenGaussians \cite{wu2024opengaussian} have shown strong capabilities in open-vocabulary semantic search for 3D scenes \cite{takmaz2025search3d}, with practical applications in robotics such as navigation \cite{huang23vlmaps, lemke2024spotcompose} and grasping \cite{lerftogo2023, shen2023F3RM, zurbrugg2024icgnet}. 

Despite these advances, existing methods face limitations in scalability, particularly in large-scale, time-sensitive settings. For instance, robots performing open-vocabulary search over vast areas, enabling an LLM-based planner to query object locations beyond the line of sight.
In such a system, a drone would map the environment, for example a large logistics center or an urban area, by capturing footage that is then processed in the cloud.
The resulting language-embedded scene would then be deployed on the robot for on-demand inference.
This approach presents several challenges. First, in time-sensitive scenarios, the drone’s survey data must be processed rapidly in the cloud (e.g., within 10 minutes). Second, encoding large-scale scenes is highly memory-intensive and may exceed the capacity of a single GPU. Third, the encoded scene—or its relevant portions—must be compact enough for deployment on a robot while remaining efficient for real-time querying. Existing methods, such as LERF~\cite{lerf2023}, LangSplat~\cite{qin2023langsplat}, OpenGaussians~\cite{wu2024opengaussian}, and SAGA~\cite{cen2023saga}, are constrained by single-GPU implementations, limiting their scalability and processing speed.

\input{figures/teaser}
To overcome these challenges, we introduce a highly scalable approach to language-embedded Gaussian splatting that enhances processing speed, alleviates memory constraints, and enables deployment in resource-limited environments.
Unlike previous methods, SLAG eliminates the need for a loss function to compute per-Gaussian language embeddings. Instead, it leverages 3D Gaussian scene parameters, computing each embedding as a normalized weighted average. Our approach maps SAM~\cite{kirillov2023segany}-masked CLIP~\cite{radford2021learningtransferablevisualmodels} embeddings onto the 3D Gaussian splatting scene using a modified Gaussian splatting rasterizer. We determine and mask the contributing Gaussian weights for each training image, then compute the final per-Gaussian embeddings as a normalized weighted sum across multiple viewpoints. This design simplifies implementation, enables highly parallelized processing, and achieves language embedding performance comparable to state-of-the-art methods.

To enhance scalability and inference performance, we store semantic embeddings separately from the Gaussian splatting model. To further optimize embedding retrieval and improve usability for downstream tasks, we propose leveraging a vector database. Designed for efficient embedding vector search, vector databases provide flexible deployment options, including embedded systems, along with built-in client APIs and management tools.
However, deploying the entire vector database on a robot may be impractical due to its size and the associated query processing overhead. To address this, we introduce a partitioning mechanism that reduces both storage requirements and query latency for resource-constrained edge devices. Instead of transferring the full database, only a subset of embeddings relevant to the robot’s immediate surroundings (e.g., within a specified radius) is downloaded, ensuring efficient and scalable operation.

In our experiments, we demonstrate an 18× speedup in augmenting Gaussian Splatting scenes with language embeddings on a 16-GPU setup compared to OpenGaussian~\cite{wu2024opengaussian}, while preserving state-of-the-art accuracy. Additionally, we showcase embeddings on a large-scale scene with 38 million Gaussians and successful deployment on an Nvidia Jetson.

In summary, this paper makes the following contributions:
\begin{itemize}
\item We introduce a highly parallelizable Gaussian scene language embedding method that enables efficient multi-GPU implementation, scaling up to process individual input images in parallel.
\item We decouple the language embeddings from the Gaussian splatting model, enabling storage and processing the language embeddings outside the GPU to address memory bottlenecks.
\item We implement and benchmark a system with 16 GPUs.
\item We propose the use of a vector database for language embedding storage along with a partitioning mechanism for deployment on resource-constrained robots.
\end{itemize}

%% file: figures/teaser.tex
\begin{figure}[t]
\vspace{5pt}
\centering
\includegraphics[width=1.0\linewidth]{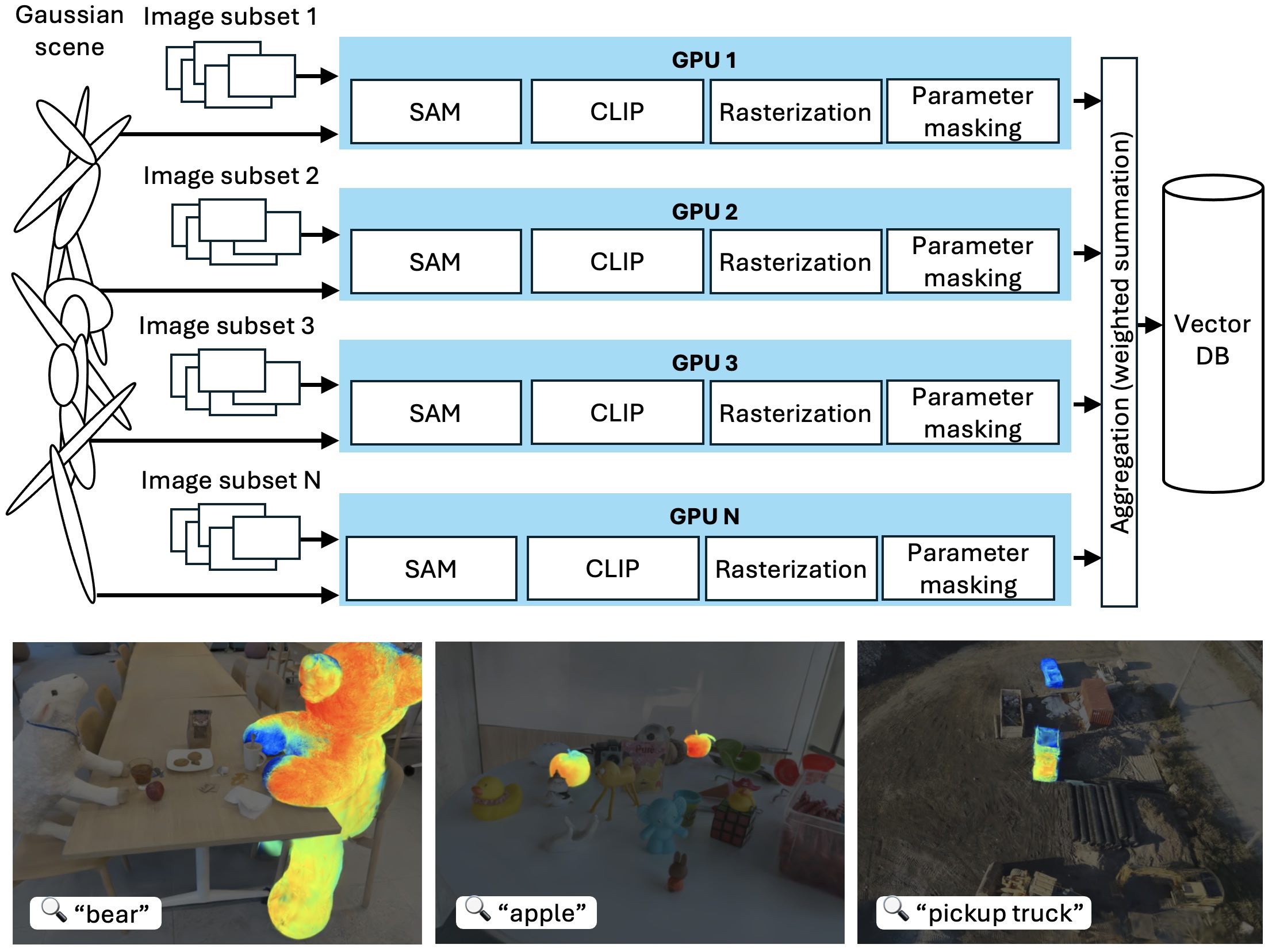}
\caption{\textbf{SLAG} is a highly scalable, multi-GPU framework for embedding language features into Gaussian splatting scene representations. It computes embeddings as normalized weighted averages of Gaussian parameters, enabling fully parallel processing (SAM-CLIP-Rasterization-Masking) with a simple aggregation step at the end. Bottom: LERF~\cite{lerf2023} teatime, figurines and MegaNerf~\cite{Turki_2022_CVPR} rubble scenes with corresponding query words and query response.}
\label{fig:teaser}
\vspace{-10px}
\end{figure}

%% file: sections/03_related_work.tex
\section{Related Work}
\label{sec:related-work}

There has been significant progress in mapping language embeddings of 2D visual-language foundation models ~\cite{radford2021learningtransferablevisualmodels} to 3D scenes representations ~\cite{lerf2023, qin2023langsplat, shi2023language,wu2024opengaussian, Weder2024labelmaker,Ji2024arkitlabelmaker, opencity3d2025, zhang2025open}. 

LERF \cite{lerf2023} integrates neural radiance fields (NeRF~\cite{mildenhall2020nerf}) with CLIP~\cite{radford2021learningtransferablevisualmodels} embeddings, creating a language embedding field by associating CLIP embeddings with 3D scene volumes across multiple scales. These volume embeddings are learned from images and their corresponding CLIP embeddings. Since CLIP embeddings are pixel-aligned, LERF employs image crops at various scales to improve localization. In contrast, our work uses Gaussian Splats~\cite{kerbl3Dgaussians}, which represent the scene with explicit 3D primitives (Gaussians) and provides greater flexibility and editability.

LangSplat \cite{qin2023langsplat} builds on LERF by replacing NeRF Gaussian Splatting for 3D scene representation.
While it also employs CLIP embeddings,
it improves pixel-alignment by leveraging object masks from the Segment Anything Model (SAM)~\cite{kirillov2023segany}.
LangSplat augments each Gaussian with 512-dimensional CLIP embeddings, learning object-specific Gaussian embeddings from masked, language-embedded images.
However, storing these high-dimensional embeddings can be memory-intensive. To address this, LangSplat employs a scene-specific autoencoder to compress embeddings. Despite its effectiveness, LangSplat has several limitations: it optimizes embeddings for 2D image planes rather than 3D spatial search; its scene-specific autoencoder introduces potential information loss in larger scenes; and it is confined to single-GPU implementations, restricting scalability and runtime.
% langsplat

LEGaussians \cite{shi2023language} similarly embeds semantic features into a Gaussian Splatting scenes but combines dense CLIP and DINO features extracted from multi-view images. This method also encounters GPU memory challenges due to the concatenated features. To alleviate this, LEGaussians introduces a quantization scheme to reduce feature size. Nevertheless, it shares limitations with LangSplat, focusing on 2D renders of Gaussian embeddings, relying on GPU-stored compressed embeddings, and being constrained to a single-GPU implementation.
% legaussian jadajada

OpenGaussian \cite{wu2024opengaussian} extends Gaussian Splatting methods from 2D pixel-level parsing to 3D point-level understanding, which is essential for robotics applications requiring precise localization and interaction.
It combines SAM masks with CLIP embeddings, mapping the SAM masks to the Gaussian scene and optimizing them using intra-mask smoothing and inter-mask contrastive losses. OpenGaussians quantizes and groups mask features before applying the final CLIP embedding.
While achieving state-of-the-art performance, its single-GPU implementation limits its scalability in large-scale or time-sensitive scenarios.
% sth about OpenGaussians

Gaussian Grouping \cite{ye2024gaussian} extends Gaussian splatting to jointly reconstruct and segment open-world 3D scenes, enabling fine-grained object-level understanding and editing. It augments each Gaussian with an Identity Encoding, grouping Gaussians into object instances or semantic categories using 2D masks generated by SAM. These identity features are trained through differentiable Gaussian rendering and optimized via spatial consistency regularization, enabling downstream tasks such as 3D object removal, inpainting, and scene recomposition.
However, its binary open-vocabulary segmentation inference requires rasterization followed by a Grounded-SAM inference step, making it impractical for resource-constrained devices. Furthermore, inference depends on a predefined camera view, requiring downstream tasks to know where to look when searching for objects, which limits its scalability in large scenes.

Segment Any GAussians (SAGA) \cite{cen2023saga} introduces a 3D promptable segmentation method that extends the Segment Anything Model (SAM) to Gaussian Splatting. By attaching scale-gated affinity features to 3D Gaussians, SAGA enables efficient multi-granularity segmentation inference within milliseconds. The method distills SAM's 2D segmentation capabilities into 3D-GS representations via scale-aware contrastive learning. To address multi-granularity ambiguity, SAGA employs a soft scale-gate mechanism to adjust feature representations based on a specified 3D physical scale. This approach facilitates real-time segmentation inference while maintaining high efficiency and scalability. SAGA supports both binary and multi-class open-vocabulary segmentation but is primarily designed for the latter, resulting in limited binary segmentation performance.

To address these challenges, we propose a method that enables true 3D point-level understanding, improves scalability through a multi-GPU implementation for substantial speedups, and mitigates GPU memory limitations by leveraging parallel processing and off-GPU embedding storage. Unlike related approaches, our method also demonstrates deployment on resource-constrained devices.

%% file: sections/04_method.tex
\section{Method}
\label{sec:method}

\begin{figure}[t]
\vspace{5pt}
     \centering
     \includegraphics[width=0.98\linewidth]{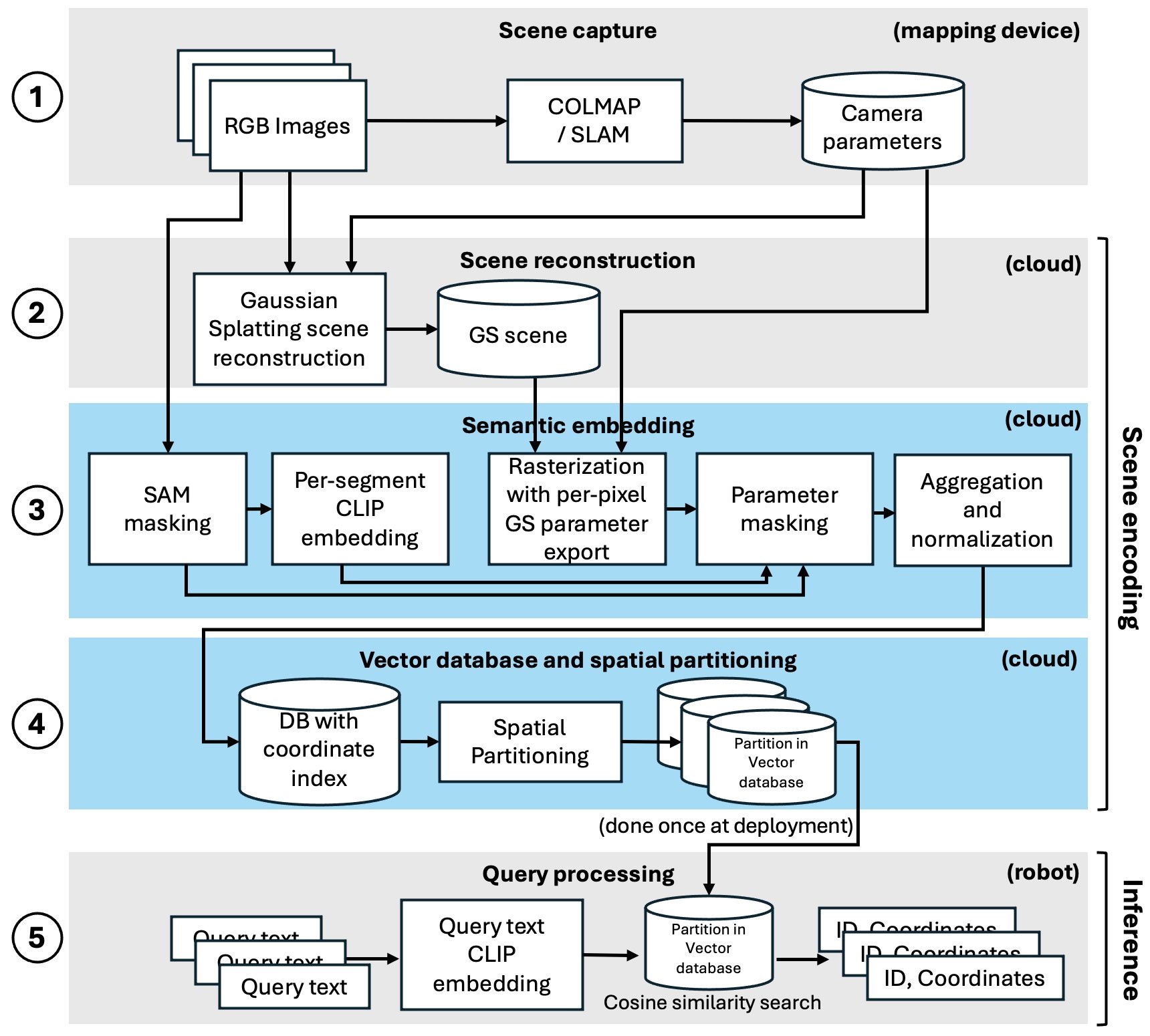}
     \caption{\textbf{Method Overview.}
     The five stages of the semantic embedding deployed across the Mapping Device, Cloud and the mobile platform (robot). This work focuses on parallelized \emph{Semantic Embedding} and \emph{Vector database}, which are highlighted in blue.}
     \label{fig:method_overview}
\vspace{-10px}
\end{figure}

Fig.~\ref{fig:method_overview} illustrates our method within an end-to-end semantic scene mapping stack. This stack consists of five key stages:

\begin{enumerate}
    \item \emph{Scene Capture} records RGB images using a camera and determines intrinsic and extrinsic camera parameters from the images via post-processing.
    \item \emph{Scene Reconstruction} builds the Gaussian splatting model from the RGB images and their intrinsic and extrinsic camera parameters.
    \item \emph{Semantic Embedding} computes language embeddings for each Gaussian.
    \item \emph{Vector Database and Partitioning} stores the embedded scene and spatially partitions it using a vector database.
    \item \emph{Query Processing} enables efficient queries on a subset of the vector database, deployed on a robot or mobile device, to retrieve coordinates for tasks such as navigation or object interaction.
\end{enumerate}

The \emph{Scene Reconstruction}, \emph{Semantic Embedding}, and \emph{Vector Database and Partitioning} stages collectively form the \emph{scene encoding phase} of the algorithm. This phase is executed once in the cloud for a given Scene Capture dataset. In contrast, the \emph{Query Processing} stage constitutes the \emph{inference phase}, which is deployed on the robot and executed repeatedly as needed for the given use case. In the following sections, we provide an overview of these stages, emphasizing our contributions: a highly parallelizable semantic embedding method, a scalable multi-GPU implementation, and the introduction of a partitioned vector database as storage.

\paragraph{Scene Capture}
This stage focuses on acquiring a sufficient number of high-quality RGB images, annotated with camera parameters (intrinsics and extrinsics), to support the 3D scene reconstruction and embedding algorithms. The most common approach for determining camera parameters is using a Structure-from-Motion algorithm, such as COLMAP \cite{schoenberger2016vote} on the RGB images. Note, that the time required to determine the camera poses can be reduced with techniques that leverage an IMU~\cite{Seiskari_2022_WACV} and/or LiDAR (e.g. Polycam 3D scanner app) hardware. This work leverages existing datasets.

\paragraph{Scene Reconstruction}
The scene reconstruction stage utilizes Gaussian Splatting \cite{kerbl3Dgaussians} due to its superior computational efficiency compared to NeRF~\cite{mildenhall2020nerf}. A key advantage of Gaussian Splatting is its explicit editability, allowing scenes to be easily partitioned into smaller components, which is important for managing large scenes. There are multi-GPU scene reconstruction implementations~\cite{zhao2024scaling3dgaussiansplatting, ye2023mathematical} that enable the creation of Gaussian splatting models with high Gaussian counts (10M+) for large scenes.

\paragraph{Semantic Embedding}
Once the Gaussian splatting scene is optimized, we compute semantic embeddings using the same training images and camera parameters from the reconstruction phase. Unlike previous methods, our approach avoids optimization-based embedding computation. Unlike previous approaches, SLAG does not rely on a loss function to compute per-Gaussian language embeddings. Instead, it leverages the 3D Gaussian scene parameters, calculating each embedding as a normalized weighted average.

\begin{figure}[H]
    \vspace{5pt}
        \centering
        \includegraphics[width=\linewidth]{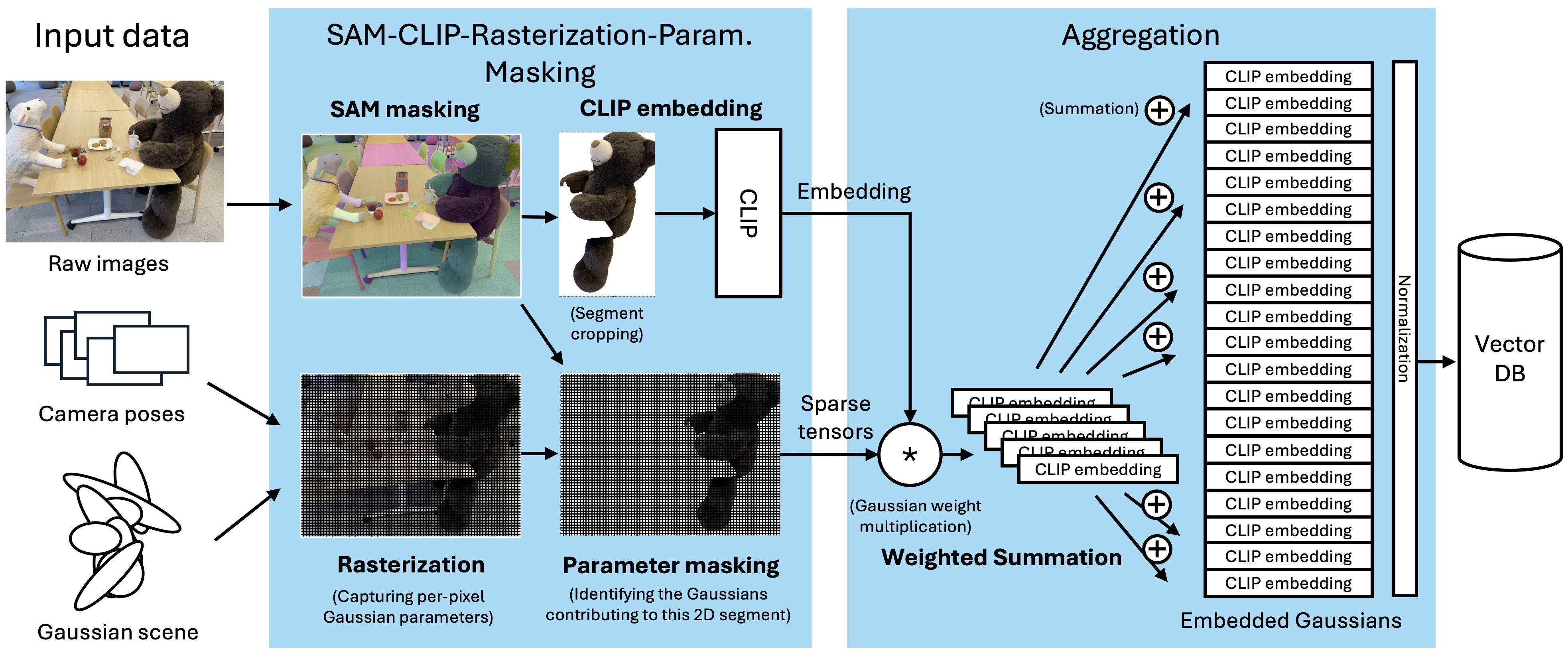}
        % \vspace{-18px}
        \caption{Embedding flow with the \textbf{SAM-CLIP-Rasterization-Parameter Masking} phase and the \textbf{Aggregation} phase}
        \label{fig:semantic_embedding_method}
        \vspace{-10px}
    \end{figure}

The semantic embedding is shown in Fig.~\ref{fig:semantic_embedding_method} for a single image. We leverage a combination of CLIP~\cite{radford2021learningtransferablevisualmodels} and SAM~\cite{kirillov2023segany} to generate object-specific embeddings in 2D. In the \textbf{SAM masking} step for each input image \( \mathbf{I}_i \) (with $i \in [1...N]$ where $N$ is the total number of posed RGB images), we produce a set of binary segmentation masks \( \mathbf{M}_j \) (with $j \in [1...M]$ where $M$ is the total number masks for the given image). In the \textbf{CLIP embedding} step we create image crops with white backgrounds for each resulting mask and generate the corresponding embeddings with CLIP
$\mathbf{E}_{ji} = \text{CLIP}(\text{crop}_j(\mathbf{I}_i \odot \mathbf{M}_j))$
where \( \mathbf{I}_i \odot \mathbf{M}_j \) is the element-wise multiplication that isolates the object within the segmentation mask and where \(\text{crop}_j(\mathbf{I}_i \odot \mathbf{M}_j)\) extracts the region of interest based on the bounding box of the mask \(M_j\).
%The bounding box cropping improves the quality of embeddings compared to using the entire masked image. 

We project 2D masked embeddings into the 3D Gaussian Splatting scene using a novel method that leverages the extracted parameters of the scene. In Gaussian splatting rendering, each Gaussian \( G_k \) in the scene contributes to the final rendered image based on its projected position and appearance parameters. Given a camera with intrinsic parameters \( \mathbf{K}_i \) and extrinsic parameters \( (\mathbf{R}_i, \mathbf{t}_i) \), the projection of a 3D Gaussian onto a 2D pixel \( p \) is determined by the rasterizer, which assigns a weight \( w_{k,p} \) representing the contribution of \( G_k \) to \( p \). The final pixel color \( C_p \) is computed as a weighted sum of the Gaussian colors \( \mathbf{c}_k \), modulated by their rasterization weights:

\begin{equation}
    C_p = \frac{\sum_k w_{k,p} \cdot \mathbf{c}_k}{\sum_k w_{k,p}}
\end{equation}
Here \( \mathbf{c}_k \) is the color of Gaussian \( G_k \), and the denominator ensures proper normalization. The weight of a Gaussian \( G_k \) at pixel \( p \) is computed as:
\begin{equation}
    w_{k,p} = \exp \left( -\frac{1}{2} (p - \mu_k')^\top \Sigma_k^{-1} (p - \mu_k') \right)
\end{equation}
Here \( \mu_k' \) is the 2D projection of the 3D Gaussian center onto the image plane and \( \Sigma_k \) is the covariance matrix defining the Gaussian’s shape in screen space.

To facilitate the projection of the 2D embeddings onto the Gaussian splatting scene we have created a modified Gaussian rasterizer to capture the $w_{k,p}$ weights of the contributing Gaussians for each pixels of rendered images:
\[
\{ w_{k,p} \}_{k \in \mathcal{K}_p} = \text{Rasterizer}_{SLAG}(G, p, \mathbf{K}_i, \mathbf{R}_i, \mathbf{t}_i)
\]
Here $G = \{ G_1, G_2, \dots, G_N \}$ represents the entire set of Gaussians in the scene, $p$ is a pixel on the rendered image $\mathbf{I}_i$, \(\mathbf{K}_i\) is the intrinsic camera matrix, \(\mathbf{R}_i\) is the rotation matrix (extrinsic parameter), \(t_i\) is the translation vector (extrinsic parameter) and $\mathcal{K}_p$ is the set of Gaussians contributing to $p$.

In the \textbf{Rasterization} step in our processing pipeline we use the $\text{Rasterizer}_{SLAG}(G, p, \mathbf{K}_i, \mathbf{R}_i, \mathbf{t}_i)$ function with the $(\mathbf{K}_i, \mathbf{R}_i, \mathbf{t}_i)$ training camera parameters corresponding to the image $\mathbf{I}_i$ used in the previous SAM masking and CLIP embedding steps.

We want to identify the Gaussians and their weights corresponding to a specific object on an image $\mathbf{I}_i$. We achieve this with masking the weights $w_{k,p}$  with the SAM masks in the \textbf{Parameter Masking} step:
\[
\{ w'_{k,p} \}_{k \in \mathcal{K}_p} = \{ w_{k,p} \cdot M_j(p) \}_{k \in \mathcal{K}_p}, \quad \forall p \in I_i
\]
Here \( M_j(p) \) is the binary SAM mask for object \( j \), where \( M_j(p) = 1 \) if pixel \( p \) belongs to the segmented object, else \( M_j(p) = 0 \), and \( w'_{k,p} \) is the masked weight, meaning only Gaussians contributing to the segmented object remain.

In the \textbf{Aggregation} step we weight the CLIP embeddings $\mathbf{E}_j$ corresponding to the masks $M_j$ with the the weights $\{w'_{k,p} \}_{{k \in \mathcal{K}_p},{p \in \mathcal{I}_i}}$ across all  training images $\mathbf{I}_i$, add them up and normalize for each Gaussian $G_k$ to determine the Gaussian embedding $\mathbf{E}_k$:

\[
\mathbf{E}_k = \frac{\sum_{i} \sum_{j} \sum_{p \in \mathbf{M}_j} w_{k, p} \cdot \mathbf{E}_j}{\sum_{i} \sum_{j} \sum_{p \in \mathbf{M}_j} w_{k, p}}
\]
The embeddings for all Gaussians are stored in a tensor:
\[
\mathbf{E}_{GAUSSIAN} = [\mathbf{E}_1, \mathbf{E}_2, \ldots, \mathbf{E}_N]
\]
where \( \mathbf{E}_{GAUSSIAN} \in \mathbb{R}^{N \times 512} \), with \( N \) being the total number of Gaussians in the scene. Inference in this system is a vector similarity search over the $\mathbf{E}_{GAUSSIAN}$ tensor.

This method eliminates the need for a cost function and iterative optimization. Note, that the SAM masking, CLIP embedding, Rasterization and Parameter Masking steps can be run independently per image and are thus highly parallelizable. The final Aggregation step is simple, and can be quickly executed, especially with further parallelization.

Unlike LangSplat~\cite{qin2023langsplat} and LEGaussians~\cite{shi2023language}, which embed semantic features directly into the Gaussian splatting model, our approach allows embeddings to be stored outside the GPU. This helps overcoming single-GPU memory limitations and avoiding the scene-specific compression of embeddings. It also enhances deployment flexibility, enabling embeddings to be stored in a dedicated database (e.g. a vector database) that improves scalability and facilitates integration with downstream tasks. Furthermore, the database can be deployed on resource-constrained devices, like a robot, without requiring a GPU for inference.

\paragraph{Scaling semantic embedding to a multi-GPU environment}

\begin{figure}[t]
\vspace{5pt}
    \centering
    \includegraphics[width=0.9\linewidth]{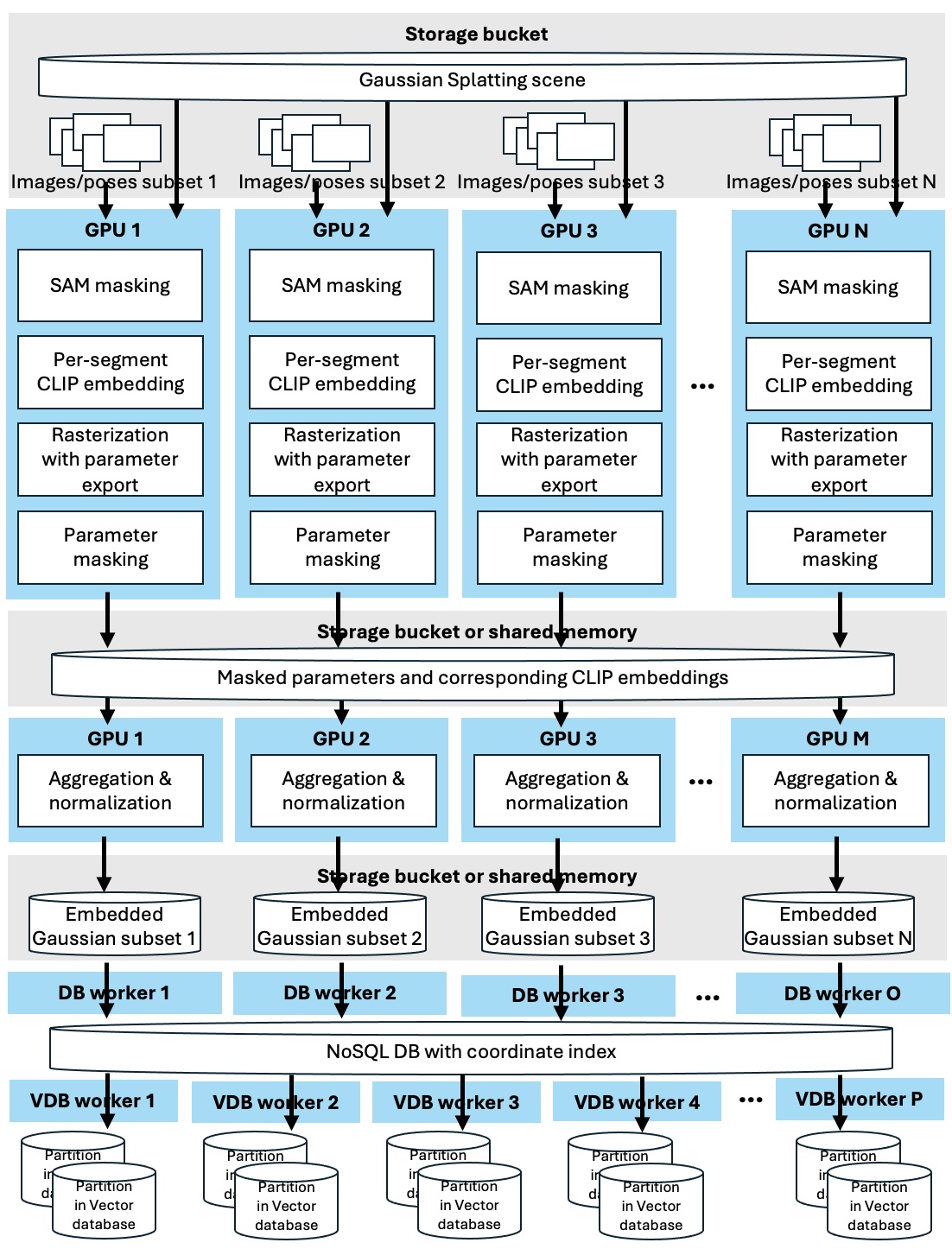}
    \caption{\textbf{Parallelization Setup} of the Semantic embedding stage \emph{(top)} and the Vector database and spatial partitioning stage \emph{(bottom)}.}
    \label{fig:parallel_method}
    \vspace{-10px}
\end{figure}

Fig.~\ref{fig:parallel_method} illustrates how our proposed method can take advantage of a multi-GPU environment.

In the \textbf{SAM-CLIP-Rasterization-Parameter masking phase}, each GPU processes a distinct subset of training images and their associated camera parameters. As discussed, the SAM masking, CLIP embedding, rasterization, and parameter masking steps are image-specific, allowing parallelization to scale with the number of available GPUs. To manage memory usage, only non-zero $w_{k,p}$ values are passed down to the aggregation phase.

In the \textbf{Aggregation phase}, for scenes, where the entire $\mathbf{E}_{GAUSSIAN}$ embedding tensor alongside the working variables fit into the GPU memory, each GPU processes a subset of the $w_{k,p}$ weights and CLIP embeddings \textit{in a single iteration}, followed by a final step that combines the results (e.g., summing tensors from 16 GPUs). For larger scenes that exceed the memory capacity of a single GPU, aggregation must be performed \textit{iteratively} over  subsets of the $\mathbf{E}_{GAUSSIAN}$ tensor and $w_{k,p}$ weights. In such cases, the final summation and normalization are also carried out in chunks, with the results concatenated outside the GPUs.

This architecture is adaptable for deployment, running efficiently on a single multi-GPU machine or scaling across multiple machines, albeit with some additional data transfer overhead.
    
\paragraph{Vector database and spatial partitioning}
To improve usability for downstream tasks and optimize embedding vector search, we propose using a vector database for storing the $\mathbf{E}_{GAUSSIAN}$ embedding tensor. Vector databases are specifically designed for efficient embedding vector retrieval, leveraging advanced indexing algorithms like Hierarchical Navigable Small World (HNSW) and Product Quantization (PQ). By indexing embedding vectors as keys, the database can perform fast similarity searches using metrics such as Euclidean distance or cosine similarity (the latter being used for CLIP~\cite{radford2021learningtransferablevisualmodels}). Using this approach, we can quickly identify query-matching Gaussians by retrieving their associated Gaussian IDs and parameters, significantly reducing the computational overhead compared to performing a brute-force similarity search over all embeddings in the scene.

For large scenes the $\mathbf{E}_{GAUSSIAN}$ embedding tensor can reach a size of tens of GBs. Storing the entire scene in a vector database on a robot is impractical due to its large size and query overhead. To address this, we pre-partition the scene and create smaller indexed vector database snapshots, allowing the robot to download only relevant subsets based on its location. We use the Gaussian mean coordinates to perform the spatial partitioning (with a NoSQL database as intermediate storage) before creating the vector database snapshots. The size of the snapshots is a design parameter and depends on the size of the area covered by the partition.
The partitioning and vector database imaging step concludes the Scene Encoding steps.

\paragraph{Query processing for inference}
For inference we only rely on the database partition deployed on the robot based on its location. The natural language queries are encoded into query vectors using the CLIP text encoder. These vectors are used to query the vector database via cosine similarity search over the $\mathbf{E}_{GAUSSIAN}$ embedding tensor partition, retrieving the parameters of matching Gaussians. A downstream task can use the returned parameters for identifying the locations of objects or visualizing them.

%% file: sections/05_experiments.tex
\section{Experiments}
\label{sec:experiments}

We first demonstrate the \textit{scene encoding speedup} achieved in a multi-GPU environment alongside qualitative results. Next, we evaluate the embedding performance through \textit{3D Binary Open-Vocabulary Segmentation} on the LERF~\cite{lerf2023} dataset, annotated by LangSplat~\cite{qin2023langsplat}, and \textit{3D Multi-Class Open-Vocabulary Segmentation} on the ScanNet~\cite{dai2017scannet} dataset, following the OpenGaussian~\cite{wu2024opengaussian} evaluation protocol. For direct comparisons, we use scenes with a moderate number of training images ($\sim$200) and moderately sized scenes ($\sim$2M Gaussians). To evaluate embedding performance on large-scale scenes, we used the Mega-NeRF~\cite{Turki_2022_CVPR} rubble scene. We only evaluated SLAG as neither of previous works could embed a scene with 38M Gaussians and 1657 training images.

\vspace{5px}\noindent\textbf{Execution time evaluation.}
We compared scene encoding times for OpenGaussian (single GPU), SLAG (single GPU), and SLAG (16 GPUs) using the same scene (scene0000\_00) from the ScanNet~\cite{dai2017scannet} dataset. The reconstructed Gaussian scene (~2M Gaussians) is small enough to fit on a single GPU, enabling direct comparison between single- and multi-GPU setups. The single-GPU tests used one NVIDIA A100 GPU with 40GB memory, while the multi-GPU setup utilized 16 NVIDIA A100 GPUs with 40GB each.
For SLAG scene reconstruction, we use the nerfstudio-splatfacto-big method with the ScanNet dataparser. The OpenGaussian Gaussian reconstruction is based on the original 3DGS method~\cite{kerbl3Dgaussians}. For straightforward comparison, we leave out the time taken for Gaussian reconstruction and we focus only on the language embedding process. Since OpenGaussian uses precomputed SAM~\cite{kirillov2023segany} masks, we added SAM masking times to their results. Both methods uses every 20th image of the ScanNet dataset (1:20 image subsampling ratio), ensuring identical SAM inference times.
The subsampling produces 252 training images. For SAM, we reuse the LangSplat~\cite{qin2023langsplat} implementation with large masks, using a $16 \times 16$ grid sampling parameter. We rasterize at $648 \times 484$ pixels resolution, while full-resolution images ($1269 \times 968$ pixels) are used for SAM masking and CLIP embedding cutouts.

\begin{figure}[t]
    \centering
    \includegraphics[width=\linewidth]{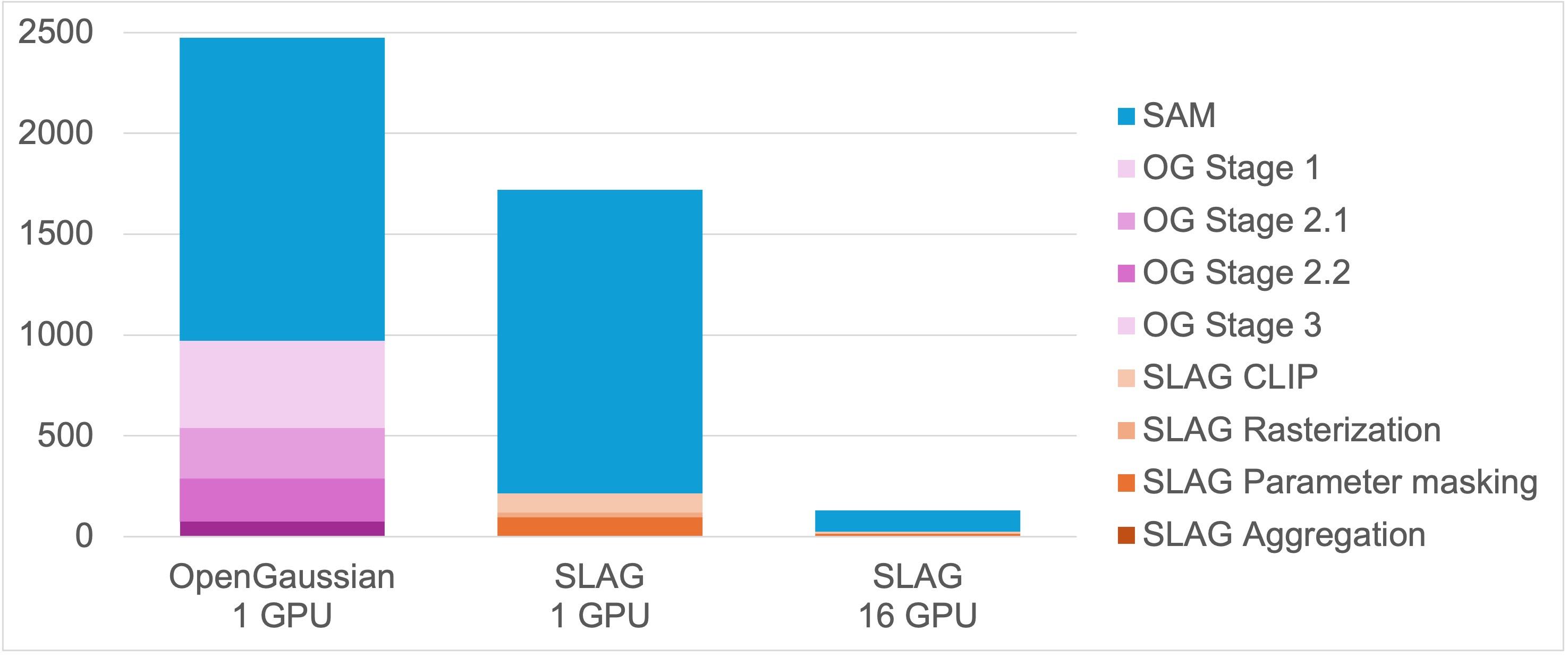}
    \vspace{-12px}
    \caption{Runtime comparison of OpenGaussians with 1 GPU, SLAG with 1 GPU and SLAG with 16 GPUs in seconds. SLAG is $1.4 \times$ and $19 \times$ faster. Units are seconds.}
    \label{fig:1vs16}
    \vspace{-10px}
\end{figure}

As shown in Fig.~\ref{fig:1vs16}, SLAG achieves a $1.4 \times$ speedup over OpenGaussian in a single-GPU configuration (over $4 \times$ if SAM is excluded).
With our 16-GPU implementation, SLAG outperforms OpenGaussian by $19 \times$, performing the embedding in 130 seconds. Notably, the SAM execution time dominates the overall runtime. As detailed in the Parameter Analysis section~\ref{subsec:parameter_analysis}, switching from the $16 \times 16$ to the $32 \times 32$ SAM grid sampling configuration, which offers slightly better performance, comes at the  cost of more than double the SAM execution time.
The relative speedups are $1.2 \times$ (single GPU) and $15 \times$ (16 GPUs) with $32 \times 32$ SAM.

\begin{figure}[t]
\vspace{5pt}
    \centering
    \includegraphics[width=\linewidth]{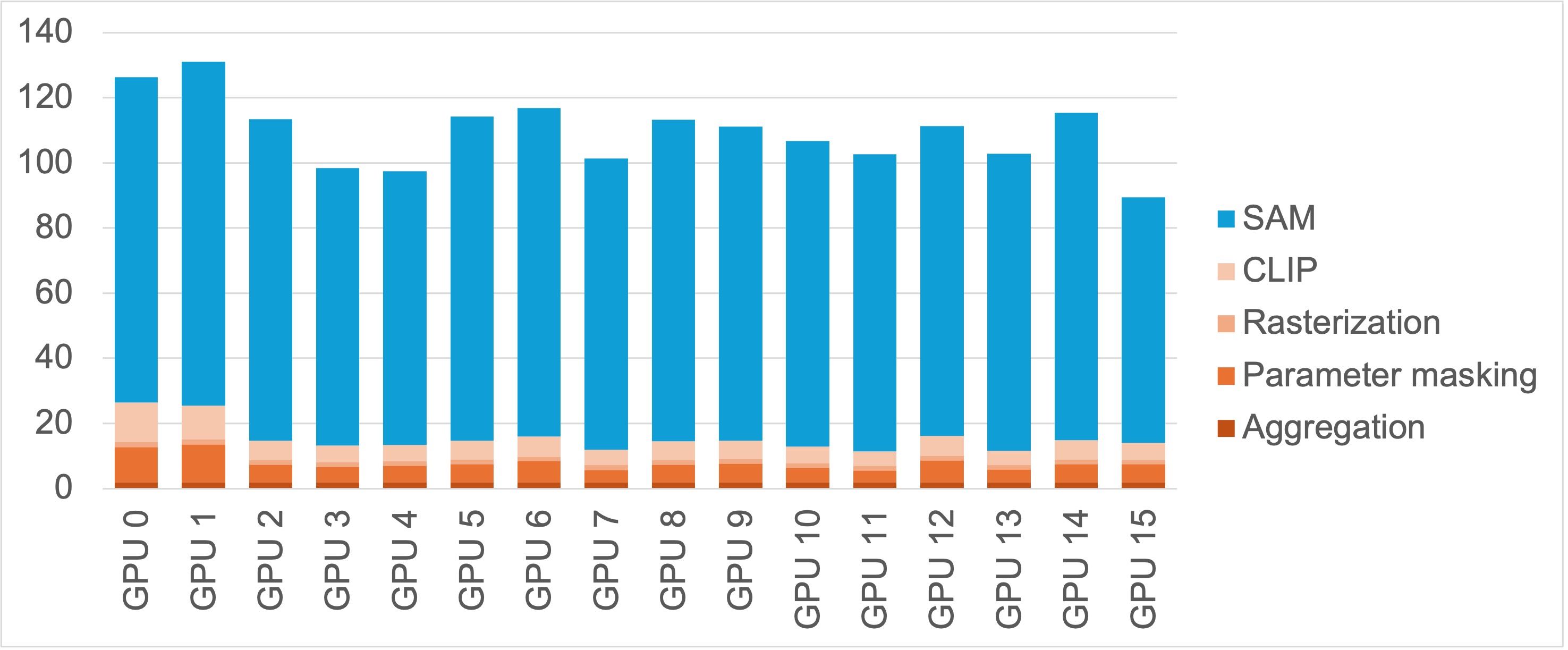}
    \vspace{-12px}
    \caption{SLAG runtime breakdown and variance across batches on 16 GPUs on ScanNet~\cite{dai2017scannet} scene0000\_00. Units are seconds.}
    \label{fig:16sn}
    \vspace{-10px}
\end{figure}

Note that the multi-GPU execution speed is determined by the GPU batch with the longest execution time. As shown in Fig.~\ref{fig:16sn}, execution times vary significantly between batches. This variation arises because the number of SAM masks varies between training images, directly impacting the number of processing operations. In our case, where images were captured sequentially and not in a randomized order, this variance is further amplified as sequential images with more segments tend to cluster within certain batches, leading to imbalanced workloads.
Also note how insignificant the Aggregation phase is.  It takes only 1 second, even if executed on a single GPU. The extent of parallelization is limited by the number of GPUs.

\vspace{5px}\noindent\textbf{Embedding large scenes.}
To evaluate embedding performance on large-scale scenes, we used the Mega-NeRF~\cite{Turki_2022_CVPR} rubble scene. We only evaluated SLAG as neither of previous works could embed such a large scene. A 38M Gaussian splatting scene was trained using the Grendel-GS~\cite{zhao2024scaling3dgaussiansplatting} multi-GPU trainer. The data set contains 1657 images; we have rasterized at a resolution of $1024 \times 768$ pixels, while a resolution of $2304 \times 1728$ pixels was used for SAM masking and CLIP embedding. The SAM-CLIP-Rasterization-Parameter masking phase ran as in the previous experiments. However, because the embedding tensor exceeded 70 GB, the Aggregation step required four iterations to process intermediate aggregation tensors. The overall embedding process took 41 minutes.

\begin{figure}[t]
    \centering
    \includegraphics[width=\linewidth]{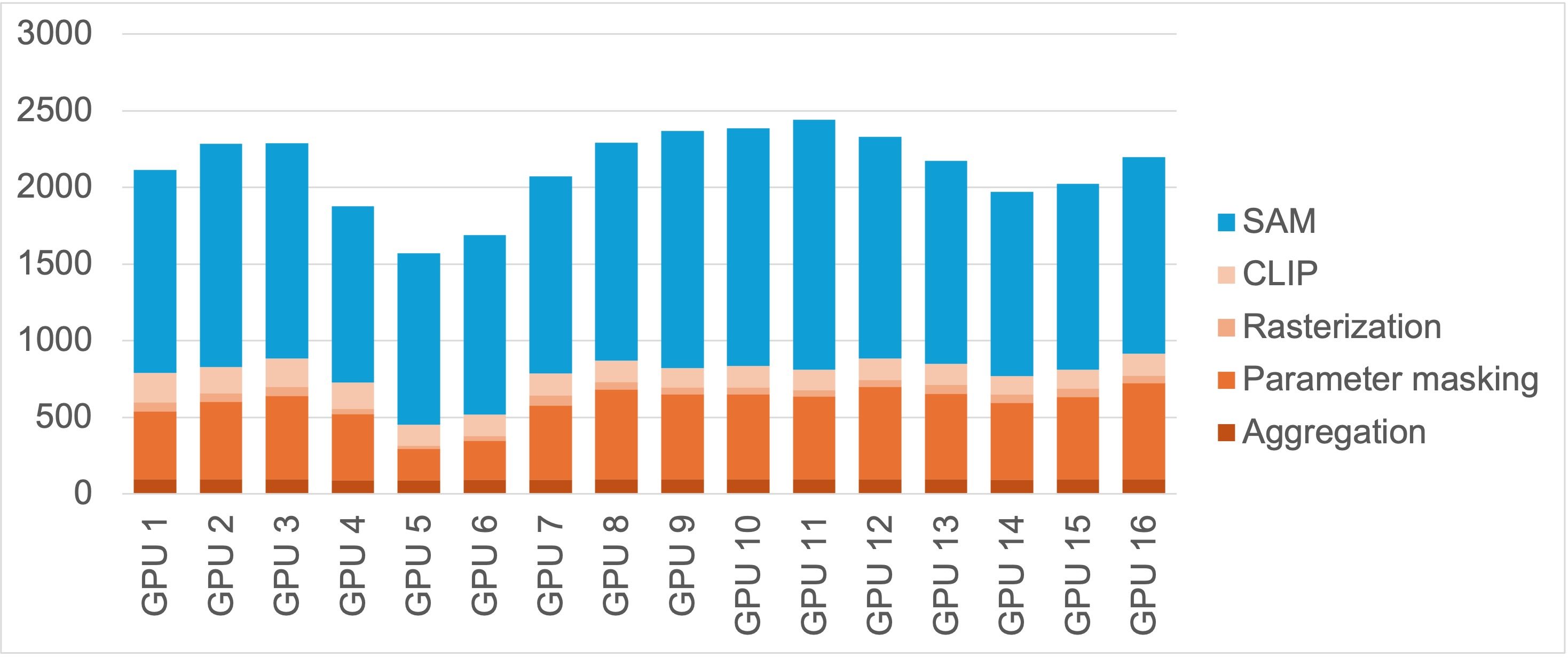}
    \vspace{-12px}
    \caption{SLAG runtime analysis across batches on 16 GPUs on Mega-Nerf~\cite{Turki_2022_CVPR} rubble scene. Units are seconds.}
    \label{fig:16rb}
        \vspace{-12px}
\end{figure}

\begin{figure}[t]
    \centering
    \includegraphics[width=\linewidth]{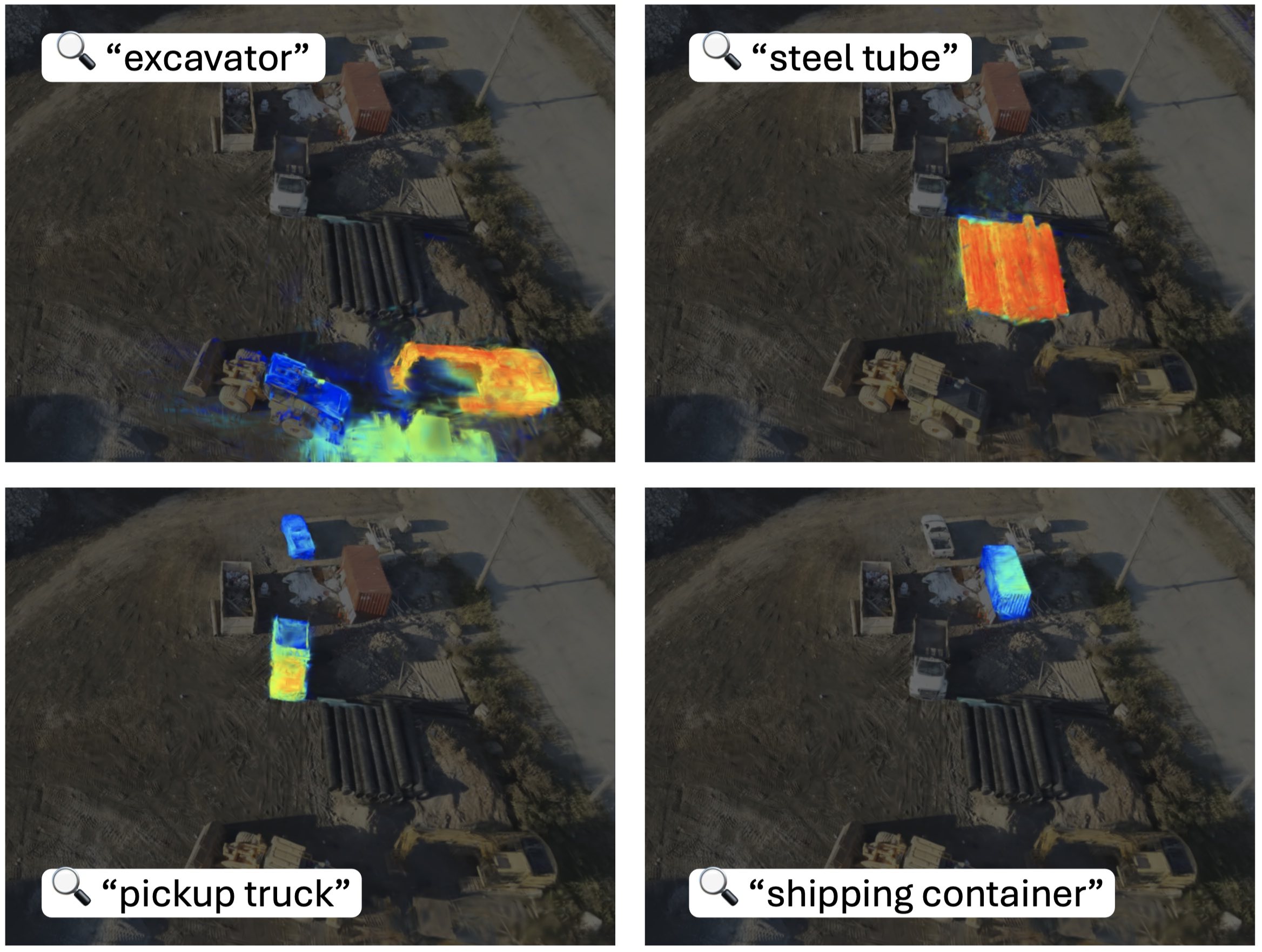}
    \caption{\textbf{SLAG qualitative results on the Mega-Nerf~\cite{Turki_2022_CVPR} rubble scene}.
    We show the natural language query in the search bar, and highlight the cosine similarity query response.}
    \label{fig:rebble_queries}
    \vspace{-15px}
\end{figure}

Fig.~\ref{fig:16rb} presents the execution time breakdown and GPU variance. The significantly larger number of Gaussians and higher rasterization resolution caused the Parameter masking and Aggregation phases to contribute more to the overall runtime. The variance in execution time was even more pronounced, consistent with the nature of the drone footage used in the dataset (and the lack of randomization).
Fig.~\ref{fig:rebble_queries} shows query results on the Mega-NeRF~\cite{Turki_2022_CVPR} rubble scene.

\vspace{5px}\noindent\textbf{Quantitative evaluation of language embeddings.}
The first evaluation task involved \textit{3D Binary Open-Vocabulary Segmentation} using the LERF~\cite{lerf2023} dataset, following the methodology outlined in OpenGaussian~\cite{wu2024opengaussian}. We evaluate four scenes—figurines, teatime, ramen, and waldo\_kitchen—using standardized viewpoints and text queries with $32\times32$ SAM and 0.28 cosine similarity. Queries are performed directly on the Gaussians in 3D space. To evaluate, we project the results to 2D using the standardized image viewpoints and compare them to the provided ground truth masks to compute mean IoU and accuracy scores.

\begin{figure}[t]
\vspace{5pt}
    \centering
    \includegraphics[width=\linewidth]{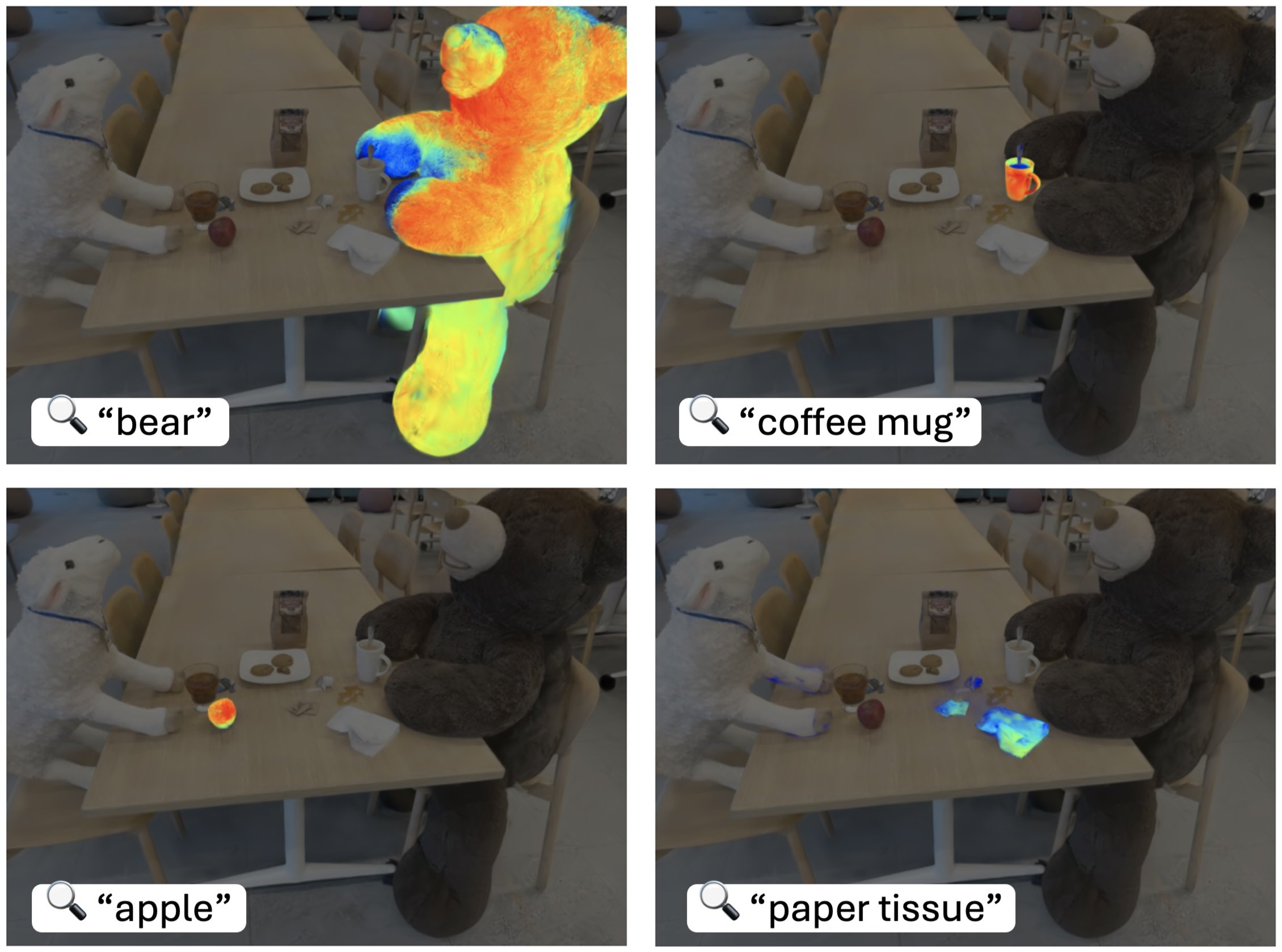}
    \caption{\textbf{SLAG qualitative results on LERF~\cite{lerf2023} teatime scene.} We show the natural language query in the search bar, and highlight the query response above the 0.28 cosine similarity threshold.}
    \label{fig:teatime_eval}
    \vspace{-20px}
\end{figure}

Table~\ref{tab:lerf_eval} presents the evaluation results, demonstrating that SLAG is on par with OpenGaussian and Gaussian Grouping. As noted by OpenGaussian~\cite{wu2024opengaussian}, LangSplat and LEGaussians are optimized for embedding projections onto a 2D plane prior to searching, which diminishes their performance when searches are conducted directly on 3D Gaussians. Fig.~\ref{fig:teatime_eval} provides qualitative examples of queries performed on the teatime scene. Note, that while SAGA performs very well on these scenes in a multi-class open-vocabulary segmentation configuration, its binary segmentation performance is poorer.

\begin{table}[t]
\caption{
\textit{3D Binary Open-Vocabulary Segmentation} scores on the LERF~\cite{lerf2023} dataset annotated by LangSplat~\cite{qin2023langsplat}.}
\centering
\resizebox{0.48\textwidth}{!}{%
\begin{tabular}{lccccc}
\toprule
%\multicolumn{6}{c}{\textbf{mIoU $\uparrow$}} \\ \hline
\textbf{Methods}                               & figurines & teatime & ramen & waldo\_kitchen & \textbf{Mean} \\ %\hline
                                               & \multicolumn{4}{c}{\textbf{mIoU $\uparrow$}} &  \\
\cmidrule(lr){2-5}

%\multicolumn{6}{c}{\textbf{mIoU $\uparrow$}} \\ \hline
 %                                              & mIoU $\uparrow$ & mIoU $\uparrow$ & mIoU $\uparrow$ & mIoU $\uparrow$ &  mean\\
LangSplat~\cite{qin2023langsplat}              & 10.16     & 11.38   & 7.92  & 9.18          & 9.66          \\
LEGaussians~\cite{shi2023language}             & 17.99     & 19.27   & 15.79 & 11.78         & 16.21         \\
SAGA~\cite{cen2023saga}                        & 15.71     & 16.65   & 19.53 & 4.54          & 14.11         \\
GaussianGrouping~\cite{ye2024gaussian}                       & \textbf{55.38}     & 33.82   & \textbf{32.58} & \textbf{30.02}            & 37.95            \\
OpenGaussian~\cite{wu2024opengaussian}         & 39.29     & \textbf{60.44}   & \underline{31.01} & 22.70  & \underline{38.36}         \\
SLAG (ours)                                 & \underline{48.1} & \underline{56.12}   & 24.76 & \underline{27.55} & \textbf{39.13} \\ %\hline
 & \multicolumn{4}{c}{\textbf{mAcc $\uparrow$}} \\ %\hline
 \cmidrule(lr){2-5}
LangSplat~\cite{qin2023langsplat}              & 8.93      & 20.34   & 11.27 & 9.09          & 12.41         \\
LEGaussians~\cite{shi2023language}             & 23.21     & 27.12   & 26.76 & 18.18         & 23.82         \\
SAGA~\cite{cen2023saga}                        & 18.59     & 17.08   & 24.4    & 5.57        & 16.41         \\
GaussianGrouping~\cite{ye2024gaussian}         & \textbf{76.61}     & 52.21   & 54.45   & \textbf{50.23}            & \underline{58.38}           \\

OpenGaussian~\cite{wu2024opengaussian}         & 55.36     & \underline{76.27}   & \underline{42.25} & 31.82         & 51.43         \\
SLAG (ours)                                 & \underline{69.95} & \textbf{81.68}  & \textbf{64.77} & \underline{48.88} & \textbf{66.32} \\
\bottomrule
\end{tabular}%
}
\vspace{-10px}
\label{tab:lerf_eval}
\end{table}

\begin{figure}[t]
\centering
\vspace{10px}
\begin{overpic}[width=0.48\textwidth]{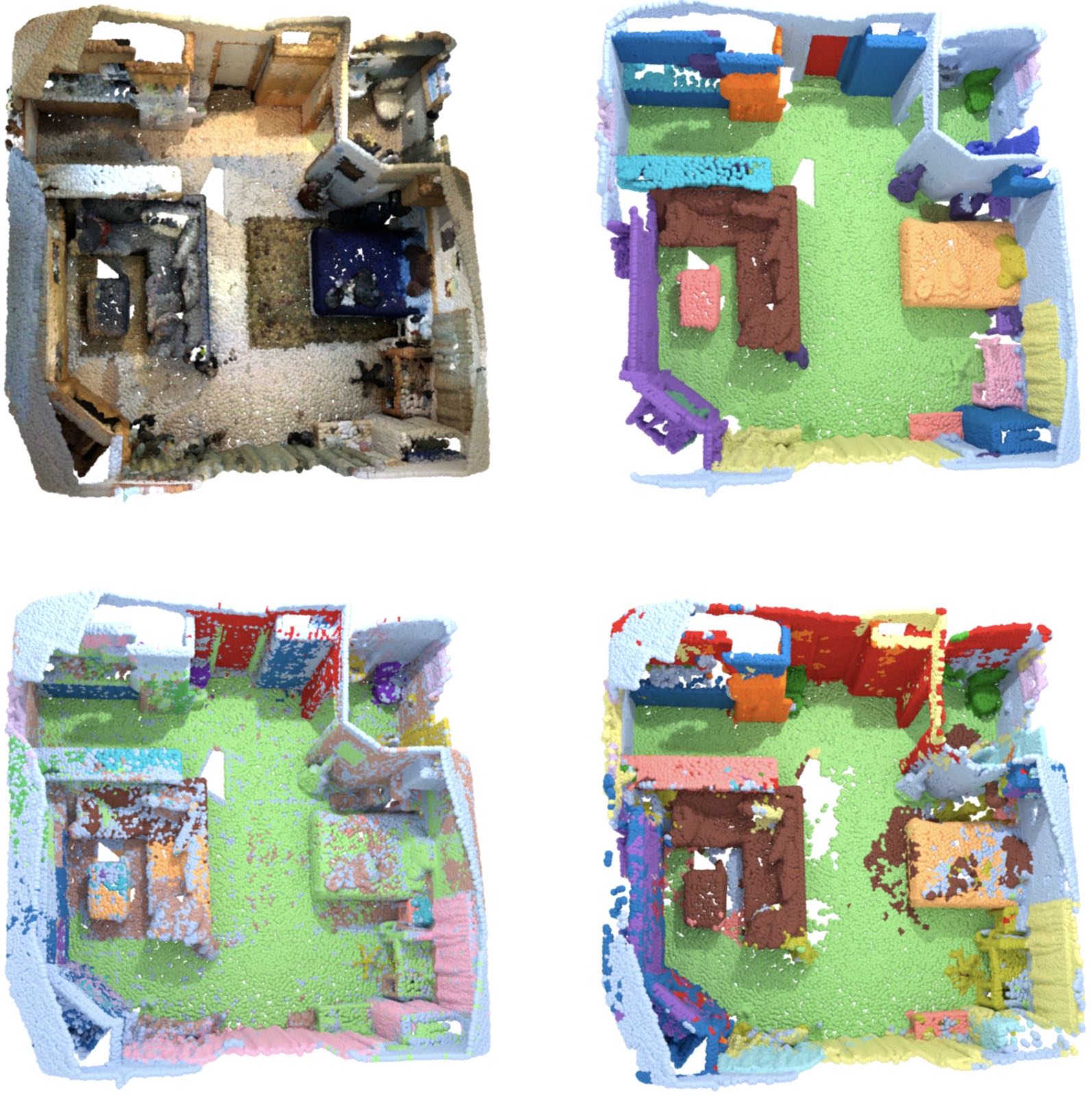}
    \put(10,101){Input Scene}
    \put(59,101){GT Semantic Masks}
    \put(7,47){OpenGaussian~\cite{wu2024opengaussian}}
    \put(65,47){SLAG (Ours)}
\end{overpic}
\vspace{-2px}
\caption{\textbf{Qualitative Segmentation Results} on ScanNet~\cite{dai2017scannet}.}
\label{fig:scene0000_eval}
    \vspace{-15px}
\end{figure}
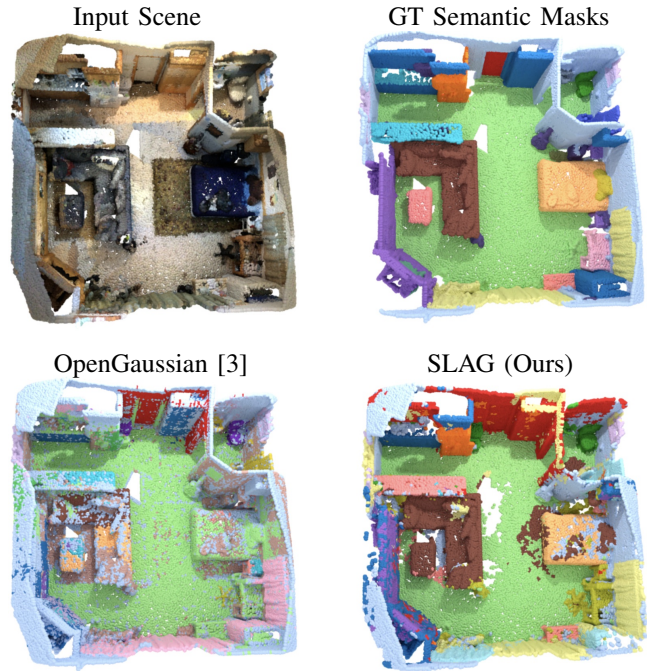

Next, we evaluate \textit{3D Multi-Class Open-Vocabulary Segmentation} on ScanNet,
following the protocol of OpenGaussian~\cite{wu2024opengaussian} and averaging results over the same 10 scenes.
Using 19 segment labels (excluding ``other furniture"), each Gaussian is assigned the class with the highest cosine similarity.
The Gaussian splatting scenes are aligned with ScanNet's point clouds, enabling Gaussian classes to be mapped to the lower-resolution point cloud vertices.
Note that for large-scale scenes like ScanNet, Gaussian Grouping is not applicable as its inference requires a viewpoint input. We also observed that SAGA's open-vocabulary variant uses an argmax operation that disproportionately favors walls and floors, degrading segmentation quality. Table~\ref{tab:scannet_eval} presents the results, showing that the SLAG baseline achieves performance comparable to OpenGaussian. Fig.~\ref{fig:scene0000_eval} provide evaluation visualizations.
 
\begin{table}[b]
\vspace{-12px}
\caption{\textbf{3D Semantic Segmentation Scores on ScanNet~\cite{dai2017scannet}}.
Scores for \cite{wu2024opengaussian, qin2023langsplat, shi2023language} are from 10 scenes as in\cite{wu2024opengaussian}. Scores for \cite{cen2023saga} are obtained using their official code. Metrics are mean intersection-over-union (mIoU) and mean accuracy (mAcc) evaluated on points.}
\centering
\scriptsize % Reduce font size
\setlength{\tabcolsep}{4pt} % Adjust column spacing
\renewcommand{\arraystretch}{1.2} % Adjust row spacing
\begin{tabular}{@{}lcccccc@{}}
\toprule
\textbf{Methods} & \multicolumn{2}{c}{\textbf{19 classes}} & \multicolumn{2}{c}{\textbf{15 classes}} & \multicolumn{2}{c}{\textbf{10 classes}} \\
\cmidrule(lr){2-3} \cmidrule(lr){4-5} \cmidrule(lr){6-7}
& \textbf{mIoU ↑} & \textbf{mAcc ↑} & \textbf{mIoU ↑} & \textbf{mAcc ↑} & \textbf{mIoU ↑} & \textbf{mAcc ↑} \\
\midrule
LangSplat~\cite{qin2023langsplat}    & 3.78  & 9.11   & 5.35  & 13.20  & 8.40  & 22.06  \\
LEGaussians~\cite{shi2023language}  & 3.84  & 10.87  & 9.01  & 22.22  & 12.82 & 28.62  \\
SAGA~\cite{cen2023saga} & 7.02 & 13.45 & 7.83 & 14.76 & 9.84 & 13.79\\
OpenGaussian~\cite{wu2024opengaussian} & \underline{24.73} & \underline{41.54}  & \textbf{30.13} & \textbf{48.25}  & \textbf{38.29} & \textbf{55.19}  \\
SLAG (ours)                   & \textbf{28.98} & \textbf{43.40}  & \underline{28.98} & \underline{44.26}  & \underline{37.14} & \underline{54.80}  \\
\midrule
SLAG PF (ours)                   & \textbf{37.54} & \textbf{49.99} & \textbf{37.71} & \textbf{50.7} & \textbf{45.46} & \textbf{60.08} \\
\bottomrule
\end{tabular}

\label{tab:scannet_eval}
    \vspace{-10px}
\end{table}

\vspace{5px}\noindent\textbf{Prediction filtering.}
To assess the potential of post-processing Gaussian embeddings, we conduct a prediction filtering evaluation. This method combines Gaussian predictions with segments produced by the ScanNet~\cite{dai2017scannet} Segmentator, which follows the graph-based algorithm from~\cite{felzenszwalb2004efficient}, assigning each segment the majority Gaussian class within it. A point cloud can be exported from the Gaussian scene for processing, and the Segmentator operates efficiently, completing segmentation in seconds for a ScanNet-sized scene. The bottom line (SLAG PF) in Table~\ref{tab:scannet_eval} demonstrate that this hybrid approach significantly improves semantic segmentation performance.

\vspace{5px}\noindent\textbf{Parameter analysis.}
\label{subsec:parameter_analysis}
The number of training images and the SAM sampling points significantly influence both prediction quality and execution time. The number of training images determines the number of SAM-CLIP-Rasterization-Weighting steps required, with a larger image set generally improving embedding quality. Similarly, increasing the SAM sampling parameter enhances segmentation quality but significantly raises computational demands, as SAM is the most resource-intensive task in our configuration. Table~\ref{tab:configs_comp} summarizes our experiments using the ScanNet~\cite{dai2017scannet} scene0000\_00 and a 16 A100 GPU setup. Increasing the SAM sampling parameter from 16x16 to 32x32 (denoted as 16 and 32 in the table) yields improved scores but results in a 2.4x increase in execution time. Similarly, reducing the training image subsampling ratio from 1:20 to 1:2 (20 and 2 in the table) further improves scores but increases embedding time by approximately 10.

\begin{table}[bt]
\vspace{1.0em}
\setlength{\tabcolsep}{3pt}
\centering
\caption{The effect of sub-sampling in the ScanNet dataset.}
\label{tab:configs_comp}
\resizebox{0.48\textwidth}{!}{%
\begin{tabular}{lccccccc}
\toprule
\textbf{Config} & \multicolumn{2}{c}{\textbf{19 classes}} & \multicolumn{2}{c}{\textbf{15 classes}} & \multicolumn{2}{c}{\textbf{10 classes}} & \textbf{Time} \\
\cmidrule(lr){2-3} \cmidrule(lr){4-5} \cmidrule(lr){6-7}
                   & \textbf{mIoU $\uparrow$} & \textbf{mAcc $\uparrow$} & \textbf{mIoU $\uparrow$} & \textbf{mAcc $\uparrow$} & \textbf{mIoU $\uparrow$} & \textbf{mAcc $\uparrow$} & \textbf{secs} \\
\midrule
16/20 & 28.67 & 48.89 & 28.22 & 47.03 & 46.15 & 72.85 & \textbf{129} \\
32/20 & 28.86 & 45.34 & 28.17 & 45.69 & 48.41 & 73.51 & 309 \\
16/2  & 31.02 & \textbf{49.93} & 30.47 & \textbf{50.41} & 47.53 & \textbf{73.49} & 1328 \\
32/2 	& \textbf{31.31} & 49.77 & \textbf{30.68} & 49.99 & \textbf{48.29} & 73.45 & 3026 \\

\bottomrule
\end{tabular}%
}
\vspace{-10px}
\end{table}

\vspace{5px}\noindent\textbf{Inference analysis.}
For inference, we deploy the $\mathbf{E}_{GAUSSIAN}$ embedding tensor on an NVIDIA Jetson Orin using a Qdrant vector database.
We evaluate partition sizes (1M-8M embeddings) and query response sizes (1k–100k Gaussians), both of which impact performance.
Table \ref{tab:database_inference} reports these results.
Response time grows with the number of retrieved Gaussians,
but typical queries (10k Gaussians) complete within tens of milliseconds, usually well under a second.
CLIP embedding adds additional 0.20s overhead.
For comparison, we also benchmark a PyTorch cosine similarity search, which uses less RAM than Qdrant, though Qdrant offers better compression for disk storage.
\begin{table}[t]
\setlength{\tabcolsep}{10pt}
\centering
\caption{Query execution time and memory footprint of the vector database deployed on a Jetson Orin depending the database size (Gaussians in millions) and the query size (number of matching Gaussians returned). The CLIP embedding is an additional 0.25 seconds. PyTorch implementation is added as reference.} 
\label{tab:database_inference}
%\resizebox{0.48\textwidth}{!}{%
\begin{tabular}{lcccc}
\toprule
\textbf{Query size} & \textbf{1M} & \textbf{2M} & \textbf{4M} & \textbf{8M} \\
\midrule
1k (s)        & 0.02	& 0.02	& 0.02	& 0.03 \\
10k (s)       & 0.09	& 0.11	& 0.12	& 0.23 \\
50k (s)      & 0.43	& 0.64	& 1.16	& 1.32 \\
100k (s)     & 0.52	& 0.74	& 1.36	& 2.37 \\
PyTorch (s)    & 0.05	& 0.09	& 0.17	& 0.33 \\
\midrule
\textbf{Storage footprint} \\
Vector DB RAM (GB)      & 2.9	& 4.9	& 9.2	& 17.7 \\
Vector DB Disk (GB)     & 0.8	& 1.3	& 2.3	& 4.3 \\
PyTorch RAM (GB)     & 2.0	& 4.0	& 8.0	& 16.0 \\
\bottomrule
\end{tabular}%
%}
    \vspace{-10px}
\end{table}

%% file: sections/06_conclusion.tex
\section{Conclusion}
\label{sec:conclusion}

This paper introduces a scalable framework for embedding semantic features into large Gaussian splatting scenes, addressing key limitations of existing approaches in speed, memory efficiency, and deployment flexibility. By leveraging multi-GPU setups and novel techniques for embedding computation and storage, our method processes large-scale scenes at unprecedented speeds while preserving high-quality results.
Our multi-GPU implementation delivers substantial performance gains, achieving an 18× speedup over single-GPU baselines and efficiently handling scenes that exceed GPU memory limits through partitioning and iterative processing. By decoupling embeddings from the Gaussian splatting model and storing them in a vector database, we facilitate seamless integration with downstream applications, including deployment on resource-constrained edge devices.

\section*{Acknowledgment}

Francis Engelmann is supported by an SNSF PostDoc.Mobility grant. This work is supported by
the National Science Foundation under Grant Number \#2342246. We thank Ruilong Li (UC Berkeley, Nerfstudio~\cite{nerfstudio} team) for support and insights on the gsplat~\cite{qin2023langsplat} implementation.